\documentclass{article}

\usepackage[preprint]{neurips_2025}

\usepackage[utf8]{inputenc} % allow utf-8 input
\usepackage[T1]{fontenc}    % use 8-bit T1 fonts
\usepackage{hyperref}       % hyperlinks
\usepackage{url}            % simple URL typesetting
\usepackage{booktabs}       % professional-quality tables
\usepackage{amsfonts}       % blackboard math symbols
\usepackage{nicefrac}       % compact symbols for 1/2, etc.
\usepackage{microtype}      % microtypography
\usepackage{xcolor}         % colors

\usepackage{amsmath,amsfonts}
\usepackage{array}
\usepackage[caption=false,font=normalsize,labelfont=sf,textfont=sf]{subfig}
\usepackage{textcomp}
\usepackage{stfloats}
\usepackage{url}
\usepackage{verbatim}
\usepackage{graphicx}
\usepackage{booktabs}
\usepackage{threeparttable}
\usepackage{multirow}
\usepackage{array}
\usepackage{pifont}
\usepackage{algorithm}
\usepackage{algpseudocode}
\usepackage{amsmath}
\usepackage{amssymb}
\usepackage{booktabs}
\usepackage{multirow}
\usepackage{colortbl}

\usepackage{caption}

\usepackage{tabularx} 
\usepackage{subcaption}
\usepackage{wrapfig} 

\usepackage{lipsum}

\title{Probability Distribution Alignment and Low-Rank Weight Decomposition for Source-Free Domain Adaptive Brain Decoding}

\author{
  Ganxi Xu\\
  Jinan University\\
  \texttt{xuganxi@outlook.com} \\
  \And
  Jinyi Long\thanks{Corresponding authors: Jinyi Long.} \\
  Jinan University\\
  \texttt{jinyil@jnu.edu.cn} \\
  \And
  Jia Zhang \\
  Jinan University \\
  \texttt{jiazhang@jnu.edu.cn} \\
}

\begin{document}

\maketitle

\begin{abstract}
Brain decoding currently faces significant challenges in individual differences, modality alignment, and high-dimensional embeddings.
To address individual differences, researchers often use source subject data, which leads to issues such as privacy leakage and heavy data storage burdens. 
In modality alignment, current works focus on aligning the softmax probability distribution but neglect the alignment of marginal probability distributions, resulting in modality misalignment.
Additionally, images and text are aligned separately with fMRI without considering the complex interplay between images and text, leading to poor image reconstruction. 
Finally, the enormous dimensionality of CLIP embeddings causes significant computational costs. 
Although the dimensionality of CLIP embeddings can be reduced by ignoring the number of patches obtained from images and the number of tokens acquired from text, this comes at the cost of a significant drop in model performance, creating a dilemma.
To overcome these limitations, we propose a source-free domain adaptation-based brain decoding framework.
Firstly, we apply source-free domain adaptation, which only acquires the source model without accessing source data during target model adaptation, to brain decoding to address cross-subject variations, privacy concerns, and the heavy burden of data storage.
Secondly, we employ maximum mean discrepancy (MMD) to align the marginal probability distributions between embeddings of different modalities.
Moreover, to accommodate the complex interplay between image and text, we concatenate the embeddings of image and text and then use singular value decomposition (SVD) to obtain a new embedding.
What's more, to achieve better image generation quality, we employ the Wasserstein distance (WD) to align the probability distributions of new embeddings.
Finally, in the target model adaptation phase of source-free domain adaptation, we employ low-rank adaptation (LoRA) to reduce the high expense of tuning the target model.
Sufficient experiments demonstrate our work outperforms state-of-the-art methods for brain decoding tasks. 
\end{abstract}

\section{Introduction}
\label{sec:introduction}
Brain decoding aims to reconstruct visual stimuli from brain signals, primarily acquired through functional magnetic resonance imaging (fMRI)~\cite{logothetis2008we}.
This approach not only advances our understanding of neural representation but also holds promise for applications in brain-computer interfaces, neurorehabilitation, and personalized cognitive assessment~\cite{wolpaw2002brain, sitaram2017closed, haynes2006decoding}.
Over the past few years, brain decoding has been able to reconstruct more realistic images with the help of increasingly powerful generative models~\cite{takagi2023high, ozcelik2023natural, wang2024mindbridge}.
However, it still faces some significant challenges.
Firstly, the brain activity exhibits high variability across subjects~\cite{chen2023seeing}.
This implies that a model trained on one subject has limited generalization when applied to another.
Secondly, fMRI data contains subjects’ personal information, and its open access poses privacy leakage risks; furthermore, brain decoding tasks require massive amounts of data, leading to significant data storage burdens.
Although Wang \emph{et al.}~\cite{wang2024mindbridge} propose a cross-subject brain decoding framework, during the adaptation phase, this framework requires the use of source subjects' data, thereby introducing not only privacy leakage risks but also increasing data storage burdens.
Thirdly, recent studies leverage deep generative models for brain decoding by aligning brain signals with vision-language models~\cite{radford2021learning}.
Current approaches for modal alignment primarily employ the SoftCLIP loss~\cite{scotti2024reconstructing}.
This loss employs softmax probability distribution generated by a robust teacher model to provide a more effective teaching signal to the student model compared to hard labels.
However, these approaches overlook directly aligning the marginal probability distributions between embeddings of different modalities, resulting in misalignment between the distributions of fMRI and image/text data, which ultimately degrades image reconstruction performance. 
Moreover, existing methods align fMRI with images and text separately without accounting for the complex interplay between image and text modalities~\cite{yang2024mma}.
For example, Wang \emph{et al.}~\cite{wang2024mindbridge} align predicted image embeddings with CLIP~\cite{radford2021learning} image embeddings and predicted text embeddings with CLIP text embeddings.
This approach fails to account for the complex interplay between images and text, leading to misaligned probability distributions across different modalities.
Finally, to ensure the quality of image reconstruction, CLIP image embeddings incorporate both category-related and patch embeddings, while CLIP text embeddings account for the number of tokens~\cite{ozcelik2023natural}.
However, the high dimensionality of these embeddings results in significant computational costs.
To address these challenges, we propose a brain decoding framework based on source-free domain adaptation (SFDA), which not only enables cross-subject and cross-modal brain decoding but also mitigates privacy concerns, reduces computational costs, and alleviates data storage burdens.
%
% The framework is divided into three phases: source model training, target model adaptation, and inference.
The framework is divided into three phases: source model training, target model adaptation.
In the source model training phase, we train a robust source model prepared for the target model adaptation. 
The source model comprises four components: an embedder, a translator, an image head, and a text head 
The image head generates predicted image embeddings to align with CLIP image embeddings, while the text head produces predicted text embeddings to align with CLIP text embeddings.
To achieve direct alignment of marginal probability distributions across modalities, we incorporate Maximum Mean Discrepancy (MMD) to perform modality alignment.
During the target model adaptation phase, we access the parameters of the source model without accessing any source subject data, thereby achieving privacy preservation and alleviating data storage burdens.
Following the source model training phase, we also utilize MMD to perform marginal distribution alignment.
What’s more, building upon the complex interplay between images and text, we concatenate the predicted image and text embeddings with the corresponding CLIP image and text embeddings, forming both the predicted and CLIP unified embeddings.
To extract meaningful features from these unified embeddings, we apply Singular Value Decomposition (SVD), where the resulting singular values are used as the new unified embeddings. 
Inspired by WGAN~\cite{arjovsky2017wasserstein}, we employ Wasserstein distance (WD) to measure the probability distributions discrepancy between the new unified embeddings, thereby achieving superior image generation performance.
Finally, we apply low-rank adaptation (LoRA) to both the image head and text head of the target model to mitigate the excessive computational demands caused by the high dimensionality of image and text embeddings, while ensuring the quality of image reconstruction.
%

%
% During the inference phase, we input the image and text data into the target model, trained during the adaptation phase, to predict the corresponding image and text embeddings. 
%
% Then, these embeddings are fed into a multi-modal versatile diffusion (VD) model~\cite{xu2023versatile} for image reconstruction.
%

%
Our main contributions are summarized as follows.
\begin{itemize}
\item We introduce source-free domain adaptation (SFDA) in brain decoding to alleviate issues of cross-subject variations, privacy concerns, and the high burden of data storage. 
\item We employ maximum mean discrepancy (MMD) to align the marginal probability distributions of fMRI with images and text, thereby addressing the incomplete alignment caused by relying solely on the SoftCLIP loss.
\item We consider the complex interplay between image and text modalities, concatenate their embeddings, and apply singular value decomposition (SVD) to obtain unified embeddings.
\item We leverage Wasserstein distance (WD) to align the probability distributions of unified embeddings, thereby significantly improving image generation quality.
\item We apply low-rank adaptation (LoRA) to our model, reducing computational costs while maintaining the model's strong learning capacity.
\item We perform extensive experiments and have shown that our method has achieved state-of-the-art performance among the existing brain decoding methods.
\end{itemize}

\section{Related Works}
\label{sec_related_works}

\subsection{Brain Decoding}
Brain decoding has attracted increasing attention from researchers due to the outstanding performance of Latent Diffusion Models in high-resolution image synthesis~\cite{rombach2022high}, the advent of multimodal models like CLIP~\cite{radford2021learning}, and the availability of new fMRI datasets~\cite{allen2022massive}.
However, most approaches~\cite{takagi2023high, ozcelik2023natural, scotti2024reconstructing} adopt a per-subject-per-model fashion, which demonstrates significantly degraded cross-subject generalization performance when applied to a new subject.
To address this limitation, Wang \emph{et al.}~\cite{wang2024mindbridge} develop a cross-subject framework for multiple subjects using only one model.
However, their approach does not actually use only one model; instead, it sets up independent embedders for each subject. 
Additionally, when adapting the model to the target subject, data from source subjects is used, which raises privacy concerns.
Our work mitigates above issues through a SFDA~\cite{liang2020we} mechanism.

\subsection{Cross-Modal Learing with Brain Signals}
Recent developments in Vision-Language Models (VLMs)~\cite{radford2021learning, jia2021scaling, yao2021filip} have greatly influenced the field of computer vision, especially in tasks that integrate language and images.
The advent of VLMs has demonstrated their potential as bridges across modalities, enhancing interactions with brain signals~\cite{takagi2023high, ozcelik2023natural}. 
To achieve better alignment between fMRI and images, Scotti \emph{et al.}~\cite{scotti2024reconstructing} design the SoftCLIP loss, where the softmax probability distribution generated by a powerful teacher model serves as a superior teaching signal for a student model compared to hard labels.
However, this loss function ignores the direct alignment of the marginal probability distributions between fMRI and other modality.
To overcome this problem, our work employs Maximum Mean Discrepancy (MMD) to align the marginal probability distributions.

\subsection{Source-free Domain Adaptation (SFDA)}
Due to the excellent performance of SFDA in protecting data privacy, an increasing number of researchers have applied SFDA to multimodal models.
Yin \emph{et al.}~\cite{yin2023crossmatch} apply SFDA to multimodal semantic segmentation.
Tang \emph{et al.}~\cite{tang2024source} utilize multimodal foundation models in SFDA.
Huang \emph{et al.}~\cite{huang2022relative} use SFDA to build a multimodal video classification model.
Inspired by prior works, we apply SFDA to brain decoding to mitigate cross-subject variations, privacy concerns, and data storage burdens. 

\subsection{Low-Rank Adaptation}
Low-Rank Adaptation (LoRA) is notable for using low-rank matrices to approximate weight changes during fine-tuning and allowing for seamless integration with pre-trained weights before inference, all without introducing additional computational overhead.
Due to LoRA's superior performance, it has been applied in many areas.
For example, Hyeon \emph{et al.}~\cite{hyeon2021fedpara} concentrate on the low-rank Hadamard product for federated learning.
Qiu \emph{et al.}~\cite{qiu2023controlling} introduce orthogonal fine-tuning to adapt text-to-image diffusion models for downstream tasks.
Yeh \emph{et al.}~\cite{yeh2023navigating} introduce an open-source library that offers
a wide selection of fine-tuning methodologies for Stable Diffusion.
In brain decoding, the high-dimensional embeddings lead to prohibitive computational overhead. 
To address this challenge, we employ DoRA (Weight-Decomposed Low-Rank Adaptation)~\cite{liu2024dora} to significantly reduce computational costs while maintaining image reconstruction performance.

\section{Methodology}
\label{sec_methodology}

\subsection{Problem Definition}
In this section, we detail the brain decoding task.
Specifically, the 1D fMRI voxels from the source subject are represented as $v_{s}\in \mathbb{R}^{F}$, where $F$ denotes fMRI voxel's size.
The image stimulus $I_{s}$ and image caption $C_{s}$ corresponding to fMRI voxels $v_{s}$ are encoded into image embedding $e_{s,I}$ and text embedding $e_{s,C}$ via a pretrained CLIP model.
By replacing $s$ with $t$, the corresponding representations for the target subject can be obtained.
%

%
% We clarify our framework in three stages: source model training, target model adaptation and inference.
We clarify our framework in two stages: source model training, target model adaptation.
In the source model training phase, we project the source fMRI voxels $v_{s}$ into an embedding $e_{s} = \mathcal{E}_{s}(v_{s})$ through an embedder $\mathcal{E}_{s}$.
Subsequently, the embedding $e_{s}$ is translated through a translator $\mathcal{T}_{s}$ into two distinct embeddings, $(\hat{e}_{s,I}$, $\hat{e}_{s,C}) = \mathcal{T}_{s}(e_{s})$.
% , which respectively denote the predicted CLIP image embedding and text embedding.
%
To achieve modality alignment with the CLIP-generated embeddings $e_{s,I}$ and $e_{s,C}$, the embeddings $\hat{e}_{s,I}$ and $\hat{e}_{s,C}$ are projected through an image head $\mathcal{H}_{s,I}$ and a text head $\mathcal{H}_{s,C}$, generating embeddings $\hat{e}^{\prime}_{s,I} = \mathcal{H}_{s,I}(\hat{e}_{s,I})$ and $\hat{e}^{\prime}_{s,C} = \mathcal{H}_{s,C}(\hat{e}_{s,C})$, which share the same embedding spaces as the CLIP embeddings $e_{s,I}$ and $e_{s,C}$, respectively.
Replacing the subscript $s$ with $t$ can give the process of the target model adaptation phase.
%

%
% In the inference phase, a multi-modal VD model combined with pretrained target modules integrates image stimuli and text captions as training data. 
%
% This approach predicts corresponding image and text embeddings to reconstruct images with greater semantic accuracy.
%

%
In the following content, we will discuss the source model training and target model adaptation phases separately.
Therefore, to simplify the expression, we will omit the subscripts $s$ and  $t$ that distinguish between the source model training phase and the target model adaptation phase.

\subsection{Source Model Training}
\begin{figure*}[t!]
\centering
\includegraphics[width=\linewidth]{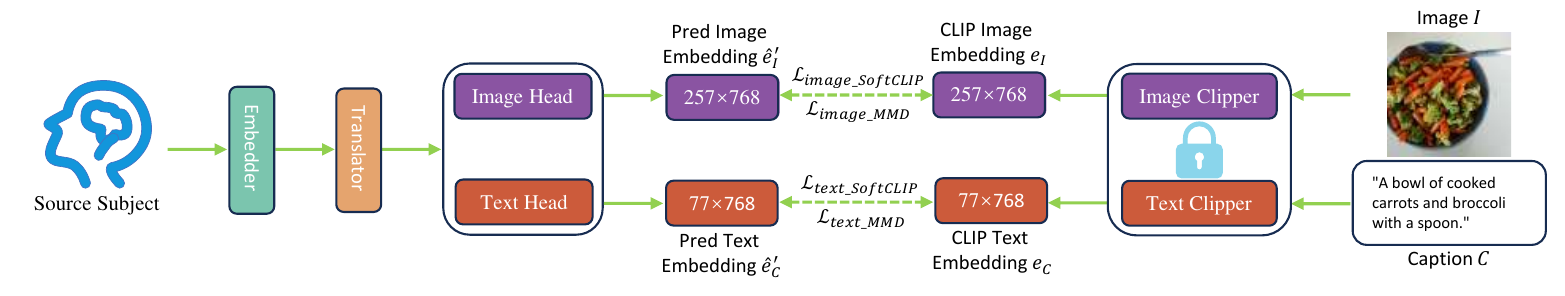}
\captionsetup{position=below, justification=raggedright, singlelinecheck=off}
\caption{
The framework of the source model training.
%
%
% During the source model training phase, we pre-train a source model using the source subject, which will be utilized in the target model adaptation phase. 
% %
% The source model consists of an embedder, a translator, an image head, and a text head. 
% %
% In this stage, we employ the SoftCLIP loss and MMD to perform modality alignment between the predicted image embeddings and CLIP image embeddings, as well as between the predicted text embeddings and CLIP text embeddings, respectively.
%
}
\label{source_model_trainig}
\end{figure*}
During the source model training phase, modality alignment is achieved using two types of loss functions, as shown in Figure~\ref{source_model_trainig}.
The first is the SoftCLIP loss~\cite{scotti2024reconstructing}, which is inspired by the concept of knowledge distillation~\cite{hinton2015distilling}. 
This loss utilizes the softmax probability distribution produced by a robust teacher model to deliver a more effective teaching signal to the student model than hard labels.
\begin{equation} 
\begin{split}
\mathcal{L}_{SoftCLIP}(\hat{e}^{\prime}, e) = -\sum^{N}_{i=1} \sum^{N}_{j=1} \Biggl[ \frac{exp(\frac{e_{i} \cdot e_{j}}{\tau})}{\sum^{N}_{m=1}exp(\frac{e_{i} \cdot e_{m}}{\tau})} \cdot log(\frac{exp(\frac{\hat{e}^{\prime}_{i} \cdot e_{j}}{\tau})}{\sum^{N}_{m=1}exp(\frac{\hat{e}^{\prime}_{i} \cdot e_{m}}{\tau})}) \Biggr],
\end{split}
\end{equation}
where $\hat{e}^{\prime}$ and $e$ represent the predicted CLIP embedding and CLIP embedding, respectively; $N$ denotes the batch size; and $\tau$ is the temperature hyperparameter.
While the SoftCLIP loss aligns predicted CLIP embeddings and CLIP embeddings in terms of their softmax probability distributions, it overlooks the direct alignment of their marginal probability distributions.
%
% Therefore, we employ maximum mean discrepancy (MMD)~\cite{gretton2012kernel} (See Appendix~\ref{sec_mmd}) to align the marginal probability distributions of the predicted CLIP embedding and CLIP embedding.
Therefore, we employ maximum mean discrepancy (MMD)~\cite{gretton2012kernel} to align the marginal probability distributions of the predicted CLIP embedding and CLIP embedding.
\begin{equation} 
\mathcal{L}_{MMD}(\hat{e}^{\prime}, e) = MMD(\hat{e}^{\prime}, e).
\end{equation}
Thus, we can derive the loss functions for the predicted image embeddings and predicted text embeddings.
\begin{align}
    \mathcal{L}_{image} &= \mathcal{L}_{SoftCLIP}(\hat{e}^{\prime}_{I}, e_{I}) + \mathcal{L}_{MMD}(\hat{e}^{\prime}_{I}, e_{I}) \label{eq_image}, \\
    \mathcal{L}_{text} &= \mathcal{L}_{SoftCLIP}(\hat{e}^{\prime}_{C}, e_{C}) + \mathcal{L}_{MMD}(\hat{e}^{\prime}_{C}, e_{C}) \label{eq_text}.
\end{align}
The overall loss function for source model training is formulated as
\begin{equation} 
\mathcal{L}_{SRC} = \mathcal{L}_{image} + \mathcal{L}_{text}.
\end{equation}

\subsection{Target Model Adaptation}
\begin{figure*}[t!]
\centering
\includegraphics[width=\linewidth]{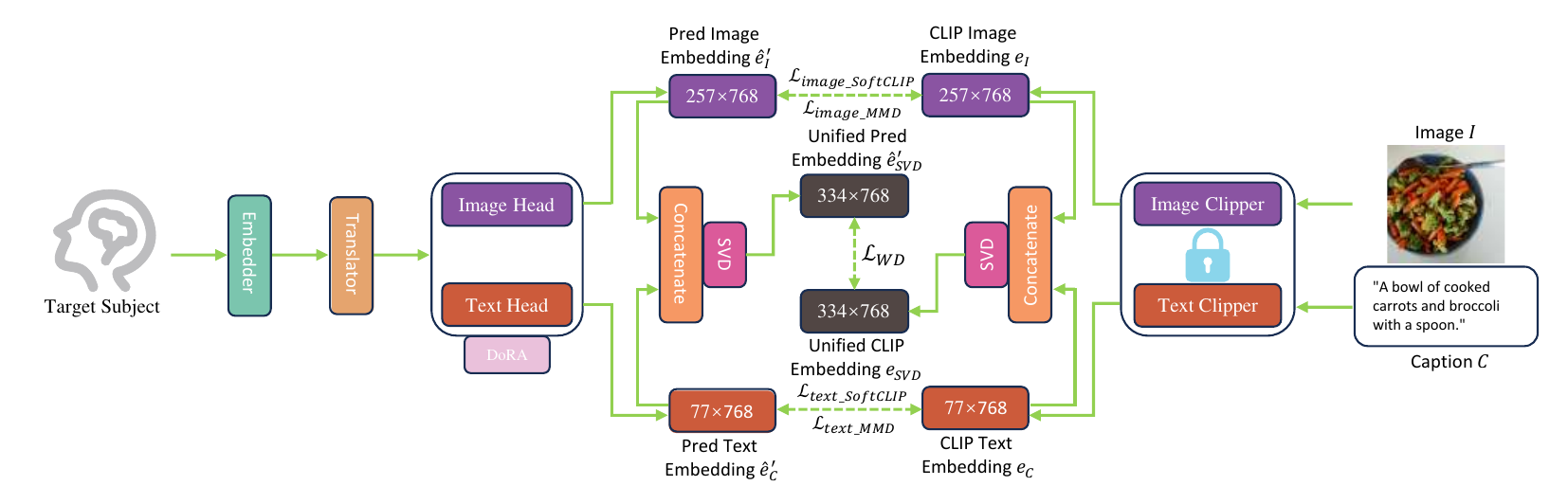}
\captionsetup{position=below, justification=raggedright, singlelinecheck=off}
\caption{
The framework of the target model adaptation.
%
% During the target model adaptation phase, we only acquire the trained model from the source model training stage without accessing the source subject's data, thereby achieving the dual objectives of protecting source subject data privacy and reducing data storage burdens. 
% %
% Similar to the source model training phase, we first employ SoftCLIP loss and MMD to perform modality alignment between the predicted image embeddings and CLIP image embeddings, as well as between predicted text embeddings and CLIP text embeddings.
% %
% Considering the complex interplay between images and text, we concatenate image embeddings and text embeddings to form a unified embedding. 
% %
% To extract more meaningful embeddings, we further process the unified embedding using SVD to get a new unified embedding. 
% %
% Subsequently, we perform modality alignment between the unified predicted embedding and unified CLIP embedding using WD.
% %
% Finally, to reduce computational overhead, we apply DoRA to both the image head and text head.
%
}
\label{target_model_adaptation}
\end{figure*}

During the target model adaptation phase, we utilize only the source model without accessing the source subject's data, thereby achieving privacy protection and reducing the data storage burden.
Similar to the source model training phase, we employ the SoftCLIP loss and MMD to perform cross-modal alignment.
\begin{align}
    \mathcal{L}_{image} &= \mathcal{L}_{SoftCLIP}(\hat{e}^{\prime}_{I}, e_{I}) + \mathcal{L}_{MMD}(\hat{e}^{\prime}_{I}, e_{I}), \\
    \mathcal{L}_{text} &= \mathcal{L}_{SoftCLIP}(\hat{e}^{\prime}_{C}, e_{C}) + \mathcal{L}_{MMD}(\hat{e}^{\prime}_{C}, e_{C}).
\end{align}
Current approaches primarily focus on aligning predicted image embeddings with CLIP image embeddings and predicted text embeddings with CLIP text embeddings separately~\cite{wang2024mindbridge,scotti2024reconstructing}, overlooking the complex interplay between images and text~\cite{yang2024mma}.
To address this limitation, we concatenate the predicted image embedding and predicted text embedding to form a unified predicted embedding, denoted as $\hat{e}^{\prime}_{u}$, as shown in Figure~\ref{target_model_adaptation}. 
Similarly, we concatenate the CLIP image embedding and CLIP text embedding to create a unified CLIP embedding, denoted as $e_{u}$.
\begin{align}
    \hat{e}^{\prime}_{u} &= Concatenate(\hat{e}^{\prime}_{I}, \hat{e}^{\prime}_{C}), \\
    e_{u} &= Concatenate(e_{I}, e_{C}).
\end{align}

To extract meaningful features from unified embeddings, we employ SVD~\cite{stewart1993early} to process the obtained unified embeddings, which requires the use of fMRI embeddings for image generation.
\begin{align}
    \hat{U}^{\prime}_{u}, \hat{\Sigma}^{\prime}_{u}, \hat{V}^{\prime}_{u} &= SVD(\hat{e}^{\prime}_{u}), \\
    U_{u}, \Sigma_{u}, V_{u} &= SVD(e_{u}), \\
    \hat{e}^{\prime}_{SVD} &= \hat{\Sigma}^{\prime}_{u}, \\
    e_{SVD} &= \Sigma_{u}.
\end{align}
The objective of brain decoding is to reconstruct visual stimuli from brain signals, thus necessitating the use of fMRI embeddings for image generation.
%
% The application of WD~\cite{ruschendorf1985wasserstein} (See Appendix~\ref{sec_wd}) in Generative Adversarial Networks (GANs)~\cite{goodfellow2020generative} for image generation offers significant inspiration for this task.
The application of WD~\cite{ruschendorf1985wasserstein} in Generative Adversarial Networks (GANs)~\cite{goodfellow2020generative} for image generation offers significant inspiration for this task.
The WD is more sensitive to local variations in distributions, enabling it to capture fine-grained differences~\cite{arjovsky2017wasserstein}.
Therefore, we employ the WD to measure the probability distribution discrepancy between the two unified embeddings.
\begin{align}
    \mathcal{L}_{WD} &= WD(\hat{e}^{\prime}_{SVD}, e_{SVD}).
\end{align}
Finally, we need to address the significant computational overhead caused by the high dimensionality of image and text embeddings.
We demonstrate this with the NSD dataset utilized in our experiments.
When performing modal alignment, we need to generate 257×768-dimensional predicted CLIP image embeddings and 77×768-dimensional predicted CLIP text embeddings from fMRI signals, and align them with CLIP image embeddings (257×768-dim extracted from the corresponding images where the first 768-dimensional vector indicates the category-related embedding and the remaining 256 embeddings represent the patches obtained from the images) and CLIP text embeddings (77×768-dim generated from the COCO captions associated with the corresponding images, where the 77 embeddings correspond to the number of tokens provided as input to the model), respectively~\cite{ozcelik2023natural}. 
As a result, the embedding dimensions become extremely large, leading to significant computational cost, as shown in Table~\ref{tab_dimensions}.
\begin{table}[htbp]
\centering
\caption{The input and output dimensions of the modules}
\label{tab_dimensions}
\begin{tabular}{ccc}
\toprule
\textbf{Module} & \textbf{Input Dimension} & \textbf{Output Dimension} \\
\midrule
Input     & $B \times 8192$ & - \\
Embedder           & $B \times 8192$ & $B \times 2048$ \\
Translator           & $B \times 2048$ & $B \times 2048$ \\
Image head           & $B \times 2048$ & $B \times 197376 (=257\times768)$ \\
Text head           & $B \times 2048$ & $B \times 59136 (=77\times768)$ \\
\bottomrule
\end{tabular}
\begin{tablenotes}
      \footnotesize
      \item[*] $B: Batch\ Size$.
    \end{tablenotes}
\end{table}
Fortunately, with the rise of large language models (LLMs)~\cite{chang2024survey}, parameter-efficient fine-tuning (PEFT)~\cite{houlsby2019parameter} has emerged, allowing fine-tuning with a small number of parameters. 
Among these methods, LoRA~\cite{hu2022lora} has gained significant popularity because it does not require changing the model architecture.
Therefore, we apply DoRA (Weight-Decomposed Low-Rank Adaptation)~\cite{liu2024dora} to the image head $\mathcal{H}_{I}$ and text head $\mathcal{H}_{C}$, which can effectively reduce computational costs while ensuring high-quality image reconstruction.
DoRA breaks down the pre-trained weights into two components: magnitude and direction. 
For fine-tuning, it specifically uses LoRA to make directional updates. 
\begin{equation} 
W^{\prime} = \underline{m} \frac{V+\Delta V}{\lVert V+\Delta V \rVert_c} = \underline{m} \frac{W_{0}+\underline{BA}}{\lVert W_{0}+\underline{BA} \rVert_c},
\end{equation}
where $m \in \mathbb{R}^{1 \times k}$ represents the magnitude vector, $\Delta V$ is the incremental directional update obtained by multiplying two low-rank matrices $B$ and $A$, and the underlined parameters indicate the trainable parameters.
Furthermore, $\lVert \cdot \rVert_{c}$ denotes the vector-wise norm of a matrix across each column.
Based on the above, we can obtain the complete loss function for the target model adaptation phase.
\begin{equation} 
\mathcal{L}_{ADP} = \mathcal{L}_{image} + \mathcal{L}_{text} + \mathcal{L}_{WD}.
\end{equation}
%

%
% The pseudocode of the target model adaptation is summarised in Appendix~\ref{sec_pseudocode}.
%

\section{Experiment}
\label{sec_experiment}

\subsection{Datasets}
Our work utilizes the Natural Scenes Dataset (NSD)~\cite{allen2022massive}, a widely-used benchmark in neuroimaging studies. 
This dataset contains high-resolution 7-Tesla fMRI recordings obtained from eight healthy adult participants during visual perception tasks involving thousands of natural images from the Common Objects in Context (COCO) dataset~\cite{lin2014microsoft}.
Consistent with established methodologies in the field, our work focuses on four participants (subj01, subj02, subj05, and subj07) who completed all scanning sessions. 
The training set for each participant comprises 8859 image stimuli and 24980 fMRI trials (with a maximum of 3 trials per image), while the test set contains 982 image stimuli and 2770 fMRI trials.
We employ preprocessed voxel data from the NSDGeneral region of interest (ROI), which varies in dimensionality across participants, containing 15724, 14278, 13039, and 12682 voxels, respectively.
The original 4D fMRI recordings (3D spatial + temporal dimensions) undergo temporal averaging and spatial vectorization prior to ROI masking, resulting in 1D voxel representations.

\subsection{Experimental Settings}
Our experimental platform consists of three Tesla V100S PCIe 32GB GPUs.
The source model training is executed for 600 epochs, while the target model adaptation phase is executed for 200 epochs. 
The batch size for both the source model training and target model adaptation phases is set to 50, the optimizer is set to AdamW, where the maximum learning rate is set to $1.5 \times 10^{-4}$.
We combine the data from three subjects as the source data, while the data from the remaining subject is used as the target data.
For example, if subj07 is selected as the target subject, the data from subj01, subj02, and subj05 are combined to form the source data.

\subsection{Results}
\begin{figure*}[htbp]
\centering
% 第1列
\begin{minipage}[t]{0.18\textwidth}
  \includegraphics[width=\linewidth]{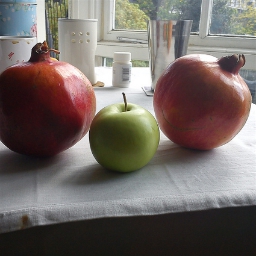} \\[2mm]
  \includegraphics[width=\linewidth]{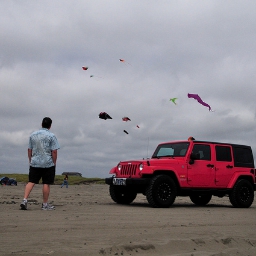} \\[2mm]
  \includegraphics[width=\linewidth]{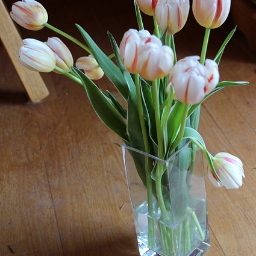}
  
  \centering
  \textbf{(A) Stimulus}
\end{minipage}
\hfill
% 第2列
\begin{minipage}[t]{0.18\textwidth}
  \includegraphics[width=\linewidth]{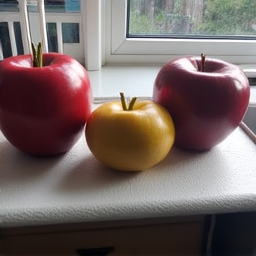} \\[2mm]
  \includegraphics[width=\linewidth]{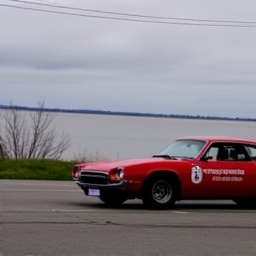} \\[2mm]
  \includegraphics[width=\linewidth]{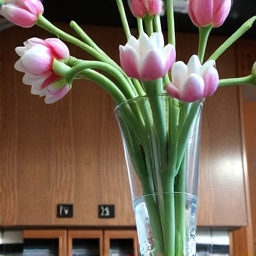}
  
  \centering
  \textbf{(B) Subj01}
\end{minipage}
\hfill
% 第3列
\begin{minipage}[t]{0.18\textwidth}
  \includegraphics[width=\linewidth]{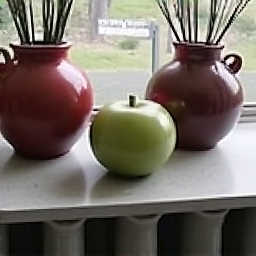} \\[2mm]
  \includegraphics[width=\linewidth]{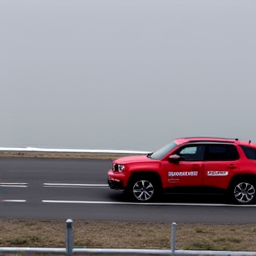} \\[2mm]
  \includegraphics[width=\linewidth]{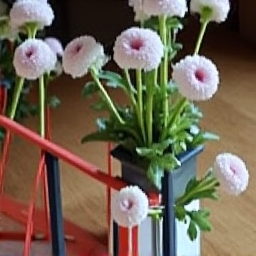}
  
  \centering
  \textbf{(C) Subj02}
\end{minipage}
\hfill
% 第4列
\begin{minipage}[t]{0.18\textwidth}
  \includegraphics[width=\linewidth]{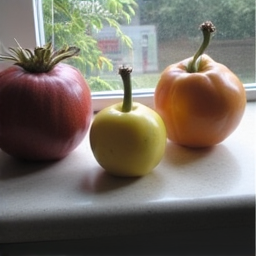} \\[2mm]
  \includegraphics[width=\linewidth]{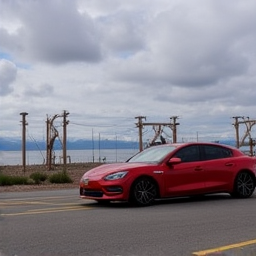} \\[2mm]
  \includegraphics[width=\linewidth]{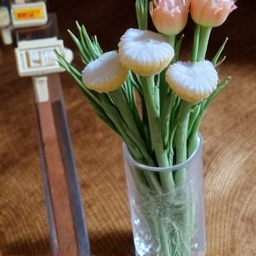}
  
  \centering
  \textbf{(D) Subj05}
\end{minipage}
\hfill
% 第5列
\begin{minipage}[t]{0.18\textwidth}
  \includegraphics[width=\linewidth]{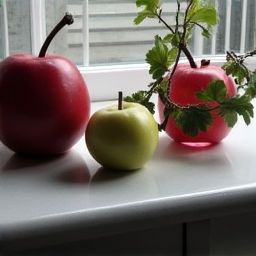} \\[2mm]
  \includegraphics[width=\linewidth]{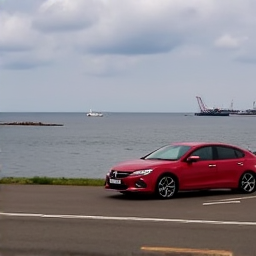} \\[2mm]
  \includegraphics[width=\linewidth]{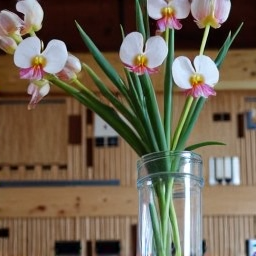}
  
  \centering
  \textbf{(E) Subj07}
\end{minipage}
\captionsetup{position=below, justification=raggedright, singlelinecheck=off}
\caption{Qualitative results.}
\label{fig_reconstruction}
\end{figure*}

\vspace{-10pt}

\begin{table*}[ht!]
\centering
\resizebox{\textwidth}{!}{
\begin{tabular}{cc|cccc|cccc}
\toprule
\multirow{2}{*}{Method} & \multirow{2}{*}{$\# \text{Models}$} & \multicolumn{4}{c|}{Low-Level} & \multicolumn{4}{c}{High-Level} \\
\cmidrule{3-6} \cmidrule{7-10} 
\multirow{2}{*}{~} & \multirow{2}{*}{~} & PixCorr $\uparrow$ & SSIM $\uparrow$ & Alex(2) $\uparrow$ & Alex(5) $\uparrow$ & Incep $\uparrow$ & CLIP $\uparrow$ & EffNet-B $\downarrow$ & SwAV $\downarrow$ \\
\midrule
Takagi \emph{et al.}~\cite{takagi2023high} & 4 & --- & --- & $83.0\%$ & $83.0\%$ & $76.0\%$ & $77.0\%$ & --- & --- \\
Brain-Diffuser~\cite{ozcelik2023natural} & 4 & 0.254 & 0.356 & $94.2\%$ & $96.2\%$ & $87.2\%$ & $91.5\%$ & 0.775 & 0.423 \\
MindEye~\cite{scotti2024reconstructing} & 4 & $\boldsymbol{0.309}$ & 0.323 & $94.7\%$ & $\boldsymbol{97.8\%}$ & $93.8\%$ & $94.1\%$ & 0.645 & 0.367 \\
DREAM~\cite{xia2024dream} & 4 & 0.288 & 0.338 & $\boldsymbol{95.0\%}$ & $97.5\%$ & $94.8\%$ & $95.2\%$ & 0.638 & 0.413  \\
MindBridge~\cite{wang2024mindbridge} & 1 & 0.151 & 0.263 & $87.7\%$ & $95.5\%$ & $92.4\%$ & $94.7\%$ & 0.712 & 0.418  \\
UMBRAE~\cite{xia2024umbrae} & 1 & 0.283 & 0.328 & $93.9\%$ & $96.7\%$ & $93.4\%$ & $94.1\%$ &	0.700 & 0.393  \\
Shen \emph{et al.}~\cite{shen2025neuro} & 1 & 0.265 & 0.357 & $93.1\%$ & $97.1\%$ & $96.8\%$ &	$\boldsymbol{97.5\%}$ & 0.633 & 0.321  \\
Our work(w/o DoRA) & 1 & 0.286 & $\boldsymbol{0.362}$ & $93.6\%$ & $97.5\%$ & $\boldsymbol{96.9\%}$ & $96.0\%$ & 0.617 & $\boldsymbol{0.311}$  \\
Our work & 1 & 0.291 & 0.355 & $93.4\%$ & $97.0\%$ & $96.8\%$ &	$96.3\%$ & $\boldsymbol{0.615}$ & 0.312  \\
\bottomrule
\end{tabular}
}
\captionsetup{position=below, justification=raggedright, singlelinecheck=off}
\caption{
Quantitative evaluation on brain decoding.
All metrics are averaged over the results from the four subjects.
}
\label{tab_results}
\end{table*}
The experimental results on several metrics (see Appendix~\ref{sec_metrics}) are shown in Table~\ref{tab_results}, with the best results highlighted in bold, while we also present the qualitative results in Figure~\ref{fig_reconstruction}.
Although three methods~\cite{wang2024mindbridge, xia2024umbrae, shen2025neuro} claim to employ only a single model in their work, they still employ separate modules for each subject during implementation.
For instance, MindBridge establishes an independent embedder for each subject.
In contrast, our approach does not introduce subject-specific modules.
We utilize only one embedder, one translator, one image head, and one text head throughout both the source model training phase and the target model adaptation phase.
The results indicate that using MMD effectively aligns fMRI with both images and text.
We concatenate the image embeddings and text embeddings and process them with SVD, which adequately considers the complex interplay between images and text. 
By using WD, we can achieve a good probabilistic alignment of the unified embeddings, thereby enhancing the quality of image generation.
In this regard, the superior performance demonstrates the effectiveness of these approaches. 
Our work introduces source-free domain adaptation into brain decoding, effectively safeguarding data privacy and alleviating the data storage burden.

\subsection{Multi-Embedder VS. Single-Embedder}
\begin{table*}[ht!]
\centering
\resizebox{\textwidth}{!}{
\begin{tabular}{ccc|cccc|cccc}
\toprule
\multirow{2}{*}{} & \multirow{2}{*}{} & \multirow{2}{*}{} & \multicolumn{4}{c|}{Low-Level} & \multicolumn{4}{c}{High-Level} \\
\cmidrule{4-7} \cmidrule{8-11} 
\multirow{2}{*}{~} & \multirow{2}{*}{~} & \multirow{2}{*}{~} & PixCorr $\uparrow$ & SSIM $\uparrow$ & Alex(2) $\uparrow$ & Alex(5) $\uparrow$ & Incep $\uparrow$ & CLIP $\uparrow$ & EffNet-B $\downarrow$ & SwAV $\downarrow$ \\
\midrule
\multicolumn{3}{c|}{Multi-Embedder} & 0.277 & 0.335 & $92.6\%$ & $95.8\%$ & $96.0\%$ & $95.7\%$ & 0.635 & 0.325  \\
\multicolumn{3}{c|}{Single-Embedder} & $\boldsymbol{0.291}$ & $\boldsymbol{0.355}$ & $\boldsymbol{93.4\%}$ & $\boldsymbol{97.0\%}$ & $\boldsymbol{96.8\%}$ &	$\boldsymbol{96.3\%}$ & $\boldsymbol{0.615}$ & $\boldsymbol{0.312}$  \\
\bottomrule
\end{tabular}
}
\captionsetup{position=below, justification=raggedright, singlelinecheck=off}
\caption{
Quantitative evaluation between multi-embedder and single-embedder.
The single-embedder corresponds to our work.
For the multi-embedder, we replicate the experimental setup of MindBridge.
}
\label{tab_multi_embedder}
\end{table*}
MindBridge~\cite{wang2024mindbridge}, UMBRAE~\cite{xia2024umbrae}, and Shen \emph{et al.}~\cite{shen2025neuro}, all claim to use only one model. 
However, each of these works sets up an independent module for subjects. 
For example, in MindBridge, an independent embedder is set up for each source subject during the training phase.
In the target model adaptation phase, not only is the embedder for each source subject retained, but an independent embedder is also set up for the target subject. 
In contrast, our method utilizes only one embedder throughout both the source model training phase and the target model adaptation phase. 
Furthermore, we do not access data from the source subjects during the target model adaptation phase.
We conduct experiments with our method following the experimental setup of MindBridge and compare it with MindBridge, as shown in Table~\ref{tab_multi_embedder}.
The table shows that our method achieves the best results across all metrics.
Moreover, our approach introduces source-free domain adaptation into brain decoding, effectively overcoming cross-subject variations, protecting data privacy, and reducing data storage burdens.

\subsection{Ablation Study on MMD and Aggregation Method}
\begin{table*}[ht!]
\centering
\resizebox{\textwidth}{!}{
\begin{tabular}{cc|cccc|cccc}
\toprule
\multirow{2}{*}{MMD} & \multirow{2}{*}{Aggregation Method} & \multicolumn{4}{c|}{Low-Level} & \multicolumn{4}{c}{High-Level} \\
\cmidrule{3-6} \cmidrule{7-10} 
\multirow{2}{*}{~} & \multirow{2}{*}{~} & PixCorr $\uparrow$ & SSIM $\uparrow$ & Alex(2) $\uparrow$ & Alex(5) $\uparrow$ & Incep $\uparrow$ & CLIP $\uparrow$ & EffNet-B $\downarrow$ & SwAV $\downarrow$ \\
\midrule
\ding{55} & \ding{55} & 0.122 & 0.242 & $85.1\%$ & $90.0\%$ & $86.6\%$ & $88.5\%$ &	0.749 &	0.443  \\
\ding{55} & \ding{51} & 0.142 & 0.257 & $87.9\%$ & $92.2\%$ & $88.3\%$ & $89.9\%$ &	0.717 &	0.392  \\
\ding{51} & \ding{55} & 0.262 & 0.354 & $92.3\%$ & $95.9\%$ & $92.5\%$ & $93.0\%$ & 0.632 & 0.346  \\
\ding{51} & \ding{51} & $\boldsymbol{0.286}$ & $\boldsymbol{0.362}$ & $\boldsymbol{93.6\%}$ & $\boldsymbol{97.5\%}$ & $\boldsymbol{96.9\%}$ & $\boldsymbol{96.0\%}$ & $\boldsymbol{0.617}$ & $\boldsymbol{0.311}$   \\
\bottomrule
\end{tabular}
}
\captionsetup{justification=raggedright, singlelinecheck=false}
\caption{Ablation study on MMD and the aggregation method.}
\label{tab_ablation_study}
\end{table*}
In this section, we study the effectiveness of MMD and the aggregation method.
In our work, to account for the complex interplay between images and text, we first concatenate the image embeddings and text embeddings.
Then, to extract more meaningful embeddings, we apply SVD to the resulting unified embeddings and use the obtained singular values as the new unified embeddings. 
Afterwards, to achieve higher-quality image generation, we utilize WD to probabilistically align the unified predicted embeddings with the unified CLIP embeddings. 
As this method involves multiple steps, we call it the aggregation method.
The experimental results are shown in Table~\ref{tab_ablation_study}.
From the table, it can be seen that when we use MMD, the performance improves significantly.
Although the SoftCLIP loss can provide an effective teaching signal through the softmax probability distribution, it still cannot guarantee the authenticity of the learned predicted embeddings~\cite{wang2024mindbridge}. 
Therefore, we use MMD to ensure more accurate predicted embeddings by aligning marginal probability distributions, resulting in notable performance improvements.
Table~\ref{tab_ablation_study} shows that the aggregation method further improves performance.
We aim to account for the complex interplay between image and text embeddings without simply concatenating them.
Therefore, we use SVD to obtain new unified embeddings. 
To enhance the quality of image generation, we use WD to align the probability distributions between the unified predicted embeddings and the unified CLIP embeddings. 
The performance improvement demonstrates the effectiveness of the aggregation method.

\subsection{Ablation Study on DoRA and LoRA}
\begin{table*}[ht!]
\centering
\resizebox{\textwidth}{!}{
\begin{tabular}{cc|cccc|cccc}
\toprule
\multirow{2}{*}{PEFT} & \multirow{2}{*}{$\#\text{Params}(\%)$} & \multicolumn{4}{c|}{Low-Level} & \multicolumn{4}{c}{High-Level} \\
\cmidrule{3-6} \cmidrule{7-10} 
\multirow{2}{*}{~} & \multirow{2}{*}{~} & PixCorr $\uparrow$ & SSIM $\uparrow$ & Alex(2) $\uparrow$ & Alex(5) $\uparrow$ & Incep $\uparrow$ & CLIP $\uparrow$ & EffNet-B $\downarrow$ & SwAV $\downarrow$ \\
\midrule
Our work(Full Fine-tuning) & 100.00 & 0.286 & 0.362 & $93.6\%$ & $97.5\%$ & $96.9\%$ & $96.0\%$ & 0.617 & 0.311   \\
\midrule
DoRA(r=4) & 6.18 & 0.258 & 0.315 & $92.0\%$ & $95.0\%$ & $93.1\%$ & $93.8\%$ & 0.724 & 0.403  \\
DoRA(r=8) & 6.34 & 0.291 & 0.355 & $93.4\%$ & $97.0\%$ & $96.8\%$ &	$96.3\%$ & 0.615 & 0.312 \\
DoRA(r=16) & 6.67 & 0.275 & 0.335 & $93.0\%$ & $96.5\%$ & $95.9\%$ & $96.0\%$ & 0.628 & 0.330  \\
DoRA(r=32) & 7.31 & 0.280 & 0.336 & $93.2\%$ & $96.8\%$ & $96.0\%$ & $96.3\%$ & 0.620 &	0.321 \\
DoRA(r=64) & 8.57 & 0.266 & 0.321 & $92.6\%$ & $95.7\%$ & $93.7\%$ & $94.0\%$ & 0.715 &	0.390 \\
\midrule
LoRA(r=4) & 6.14 & 0.175 & 0.248 & $87.4\%$ & $89.6\%$ & $86.8\%$ & $86.3\%$ & 0.812 & 0.498  \\
LoRA(r=8) & 6.30 & 0.198 & 0.274 & $89.4\%$ & $91.2\%$ & $87.6\%$ & $88.9\%$ & 0.796 & 0.472  \\
LoRA(r=16) & 6.63 & 0.246 & 0.310 & $92.8\%$ & $94.6\%$ & $92.3\%$ & $92.6\%$ & 0.747 & 0.415  \\
LoRA(r=32) & 7.27 & 0.251 & 0.313 & $92.0\%$ & $94.7\%$ & $92.5\%$ & $93.1\%$ &	0.741 &	0.410  \\
LoRA(r=64) & 8.53 & 0.227 & 0.292 & $91.2\%$ & $92.6\%$ & $89.8\%$ & $90.2\%$ &	0.772 &	0.445  \\
\bottomrule
\end{tabular}
}
\captionsetup{justification=raggedright, singlelinecheck=false}
\caption{Ablation study on DoRA and LoRA.}
\label{tab_lora}
\end{table*}
In this section, we explore the impact of different rank $r$ on DoRA and LoRA.
We assess the performance of our method when the rank $r$ is set to 4, 8, 16, 32, and 64. 
The results are shown in Table~\ref{tab_lora}.
As we can see, DoRA outperforms LoRA across all values of $r$.
Table~\ref{tab_lora} also provides the number of trainable parameters in our model for different values of $r$.
The application of DoRA and LoRA significantly reduces the number of trainable parameters.
DoRA decomposes the pre-trained weights into two components: magnitude and direction.
During fine-tuning, it employs LoRA for directional updates, effectively reducing the number of trainable parameters.
In addition, DoRA enhances the model's learning capacity, further validating the effectiveness of our approach.

\subsection{Cross-subject Variation Overcoming}

\begin{wrapfigure}{r}{0.4\textwidth}
  \centering
  \includegraphics[width=0.9\linewidth]{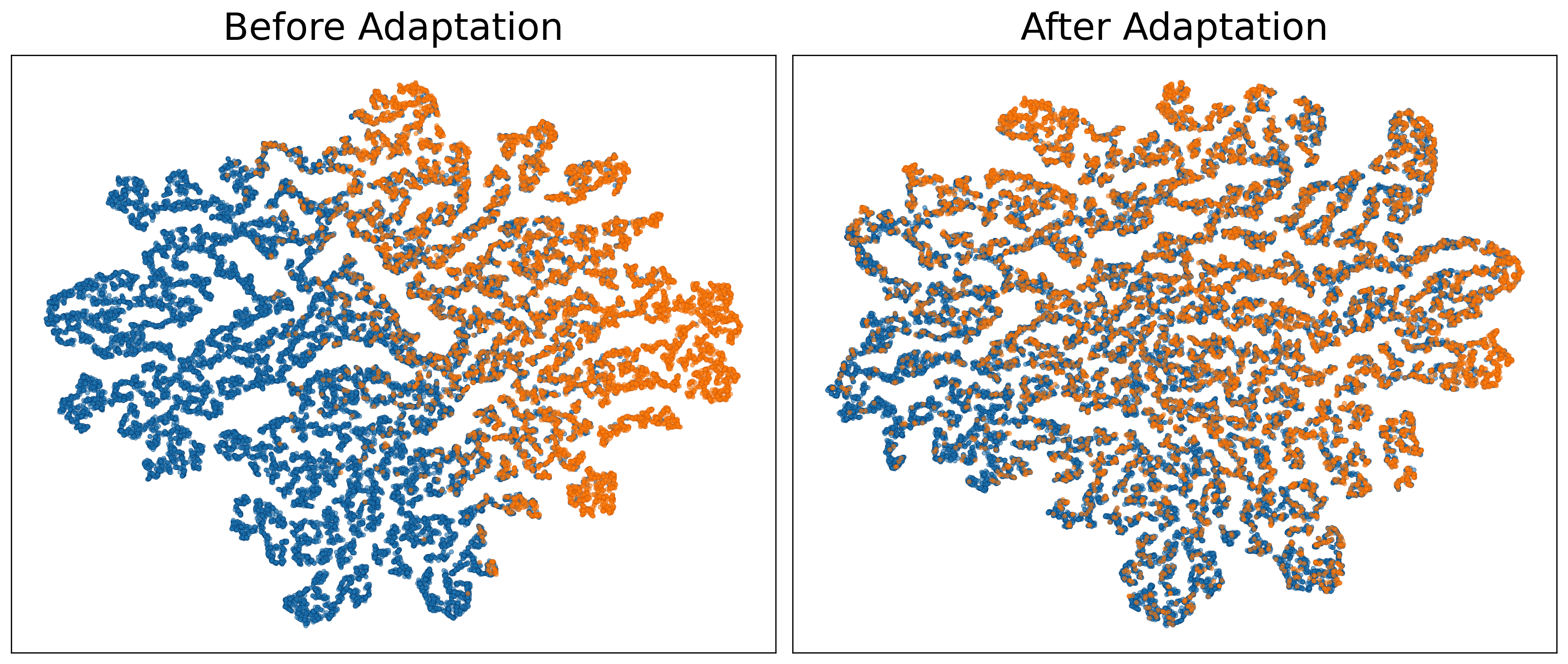}
  \caption{
  t-SNE Visualization~\cite{van2008visualizing}. 
  Let's take the case where the target subject is subj01 as an example.
  }
  \label{fig_t_sne}
\end{wrapfigure}

% \lipsum[1-3] %
%
Due to the strong variability in brain signals across different subjects, we leverage source-free domain adaptation to overcome cross-subject variation, as shown in Figure~\ref{fig_t_sne}.
It can be observed that before the target model adaptation phase, source features and target features are not mixed and can be easily distinguished. 
After the target model adaptation phase, source features and target features are mixed together and cannot be easily differentiated. 
This demonstrates that we have successfully extracted subject-invariant representations, effectively overcoming cross-subject variation.
Furthermore, when training the target model, we do not need to use source subject data, which not only protects the data privacy of the source subject but also reduces the burden of data storage.

\section{Conclusion}
\label{sec_conclusion}
In this paper, we introduce source-free domain adaptation to brain decoding to address the problem of high variability in brain signals. 
Additionally, since source-free domain adaptation does not rely on source subject data during the target model adaptation phase, it alleviates concerns about privacy leakage and the heavy burden of data storage.
Specifically, in the source model training phase, we use MMD to align the marginal probability distributions of fMRI signals with images and text, achieving modality alignment between fMRI signals and images, as well as between fMRI signals and text.
In the target model adaptation phase, we also use MMD to achieve modality alignment, similar to the source model training phase.
To account for the complex interplay between images and text, our method concatenates images and text and uses SVD to extract more meaningful embeddings. 
To achieve higher-quality image generation, we use the WD to align the probability distribution of the unified embeddings. 
Finally, to reduce the computational burden caused by the high dimensionality of the embeddings while maintaining image reconstruction performance, we leverage DoRA to significantly reduce the number of trainable parameters.
We also demonstrate the effectiveness of our method through multiple experiments.

\bibliographystyle{unsrt}
\bibliography{references}

%%%%%%%%%%%%%%%%%%%%%%%%%%%%%%%%%%%%%%%%%%%%%%%%%%%%%%%%%%%%

\appendix

\section{Technical Appendices and Supplementary Material}

\subsection{Limitations}
\label{sec_limitations}
Due to the limited availability of high-quality fMRI multimodal datasets, we conduct our experiments solely on the NSD dataset. 
The NSD dataset contains both image and text data, making it one of the rare fMRI datasets that include both modalities. 
Furthermore, the NSD data is collected using a 7T MRI system, which represents the highest precision currently used in clinical practice. 
Therefore, we believe that using the NSD dataset can still provide us with highly valuable insights. 
In the future, as more high-quality fMRI datasets become available, we will evaluate our methods on a broader range of such datasets.

\subsection{Implementation Details}
\label{sec_implementation_details}

\subsubsection{Metrics}
\label{sec_metrics}
To compare with other methods, we employed eight image quality evaluation metrics.
Pixel-level correlation (PixCorr), Structural Similarity Index (SSIM)~\cite{wang2004image}, AlexNet(2), and AlexNet(5)~\cite{krizhevsky2012imagenet} are employed as evaluation metrics for low-level visual properties.
AlexNet(2) and AlexNet(5) represent the two-way identification of the 2nd and 5th feature layers of AlexNet, respectively.
We employ Inception~\cite{szegedy2016rethinking}, CLIP, EffNet-B~\cite{tan1905rethinking}, and SwAV~\cite{caron2020unsupervised} to evaluate high-level properties.
EffNet-B and SwAV are distance metrics derived from EfficientNet-B13 and SwAV-ResNet50, respectively.

\subsubsection{Architecture details}
\label{sec_architecture_details}
Given the natural variability in brain size and structure, the fMRI signals within the ROI show variations in size.
The ROI dimensions of subj01, subj02, subj05, and subj07 are 15724, 14278, 13039, and 12682 voxels, respectively.
To standardize the input dimensions, we use Adaptive Max Pooling~\cite{wang2024mindbridge} to rescale the dimensions of the fMRI signals to 8192 voxels before feeding the data into the embedder.
The embedder, translator, image head, and text head adopt the same neural network architecture as MindBridge~\cite{wang2024mindbridge}.

\subsubsection{Pseudocode of Target Model Adaptation}
\label{sec_pseudocode}
\begin{algorithm*}
\caption{Target Model Adaptation}
\label{alg_target_model_adaptation}
\begin{algorithmic}[1]

\State \textbf{Input}: $(\text{Image},\ \text{Text},\ \text{fMRI})$
\State \textbf{Model}: $\text{Image Clipper}, \ \text{Text Clipper},\ \text{Embedder}\ \mathcal{E}, \ \text{Translator}\ \mathcal{T},\ \text{Image Head with DoRA}\ \mathcal{H}_{I},$
\State \hspace{3em} $\text{Text Head with DoRA}\ \mathcal{H}_{C}$

\Statex

\State $\#$ extract normalized embeddings from the image, text, and fMRI
\State $e_{I} = \text{Norm}(\text{Image Clipper}(\text{Image}))$
\State $e_{C} = \text{Norm}(\text{Text Clipper}(\text{Text}))$
\State $\hat{e}_{I},\ \hat{e}_{C} = \mathcal{T}(\mathcal{E}(\text{fMRI}))$
\State $\hat{e}^{\prime}_{I} = \text{Norm}(\mathcal{H}_{I}(\hat{e}_{I}))$
\State $\hat{e}^{\prime}_{C} = \text{Norm}(\mathcal{H}_{C}(\hat{e}_{C}))$

\Statex

\State $\#$ concatenate the image embeddings and text embeddings, then apply SVD to extract new unified embeddings
\State $U_{u}, {\Sigma}_{u}, V_{u} = \text{SVD}(\text{Concatenate}(e_{I},\ e_{C},\ dim=1))$
\State $\hat{U}^{\prime}_{u}, \hat{\Sigma}^{\prime}_{u}, \hat{V}^{\prime}_{u} = \text{SVD}(\text{Concatenate}(\hat{e}^{\prime}_{I},\ \hat{e}^{\prime}_{C},\ dim=1))$
\State $e_{\text{SVD}} = \text{Norm}(\Sigma_{u}$)
\State $\hat{e}^{\prime}_{\text{SVD}} = \text{Norm}(\hat{\Sigma}^{\prime}_{u}$)

\Statex
    
\State compute loss functions:
\State $\mathcal{L}_{image} = \mathcal{L}_{SoftCLIP}(\hat{e}^{\prime}_{I},\ e_{I}) + \mathcal{L}_{\text{MMD}}(\hat{e}^{\prime}_{I},\ e_{I})$
\State $\mathcal{L}_{text} = \mathcal{L}_{SoftCLIP}(\hat{e}^{\prime}_{C},\ e_{C}) + \mathcal{L}_{\text{MMD}}(\hat{e}^{\prime}_{C},\ e_{C})$
\State $\mathcal{L}_{WD} = \text{WD}(\hat{e}^{\prime}_{\text{SVD}},\ e_{\text{SVD}})$
\State $\mathcal{L}_{ADP} = \mathcal{L}_{image} + \mathcal{L}_{text} + \mathcal{L}_{\text{WD}}$

\end{algorithmic}
\end{algorithm*}

\end{document}